\documentclass[final,5p,times,twocolumn,authoryear]{elsarticle}

\usepackage{amssymb}
\usepackage{amsmath}
\usepackage{graphicx}%
\usepackage{multirow}%
\usepackage{amsmath,amssymb,amsfonts}%
\usepackage{amsthm}%
\usepackage{mathrsfs}%
\usepackage[title]{appendix}%
\usepackage{xcolor}%
\usepackage{textcomp}%
\usepackage{manyfoot}%
\usepackage{booktabs}%
\usepackage{algorithm}%
\usepackage{algorithmicx}%
\usepackage{algpseudocode}%
\usepackage{listings}%
\usepackage{subcaption}
\usepackage{tabularx}      
\usepackage{array} 
\usepackage[hyphens]{url}
\usepackage[colorlinks=true,
            linkcolor=blue,
            citecolor=blue,
            urlcolor=blue]{hyperref}
            
\newcolumntype{Y}{>{\centering\arraybackslash}X}

\journal{Engineering Applications of Artificial Intelligence}

\begin{document}

\begin{frontmatter}

\title{A practical artificial intelligence framework for legal age estimation using clavicle computed tomography scans}

%

\author[1,2,3]{Javier Venema}
\author[4]{Stefano De Luca}
\author[2,3,1]{Pablo Mesejo}
\author[5,1]{Óscar Ibáñez\corref{cor1}}
\cortext[cor1]{Corresponding author.
  E-mail address: \href{mailto:oscar.ibanez@udc.es}{oscar.ibanez@udc.es}}





\affiliation[1]{organization={Panacea Cooperative Research S.Coop.},
            addressline={C. Ramón y Cajal, 33}, 
            city={Ponferrada},
            postcode={24400}, 
            country={Spain}}

\affiliation[2]{organization={Department of Computer Science and Artificial Intelligence, University of Granada},
            addressline={C. Periodista Daniel Saucedo Aranda}, 
            city={Granada},
            postcode={18071}, 
            country={Spain}}

\affiliation[3]{organization={Andalusian Research Institute in Data Science and Computational Intelligence (DaSCI Institute)},
            addressline={Av. del Conocimiento, 37}, 
            city={Granada},
            postcode={18037}, 
            country={Spain}}

\affiliation[4]{organization={Department of Organisms and Systems Biology, Faculty of Biology, University of Oviedo},
            addressline={C. Catedrático Rodrigo Uria}, 
            city={Oviedo},
            postcode={33066}, 
            country={Spain}}

\affiliation[5]{organization={Faculty of Computer Science, CITIC, University of A Coruña},
            addressline={Campus de Elviña}, 
            city={A Coruña},
            postcode={15071}, 
            country={Spain}}

\begin{abstract}
Legal age estimation plays a critical role in forensic and medico-legal contexts, where decisions must be supported by accurate, robust, and reproducible methods with explicit uncertainty quantification. While prior artificial intelligence (AI)-based approaches have primarily focused on hand radiographs or dental imaging, clavicle computed tomography (CT) scans remain underexplored despite their documented effectiveness for legal age estimation.
In this work, we present an interpretable, multi-stage pipeline for legal age estimation from clavicle CT scans. The proposed framework combines (i) a feature-based connected-component method for automatic clavicle detection that requires minimal manual annotation, (ii) an Integrated Gradients–guided slice selection strategy used to construct the input data for a multi-slice convolutional neural network that estimates legal age, and (iii) conformal prediction intervals to support uncertainty-aware decisions in accordance with established international protocols.
The pipeline is evaluated on 1,158 full-body post-mortem CT scans from a public forensic dataset (the New Mexico Decedent Image Database). The final model achieves state-of-the-art performance with a mean absolute error (MAE) of 1.55 $\pm$ 0.16 years on a held-out test set, outperforming both human experts (MAE of approximately 1.90 years) and previous methods (MAEs above 1.75 years in our same dataset). Furthermore, conformal prediction enables configurable coverage levels aligned with forensic requirements. Attribution maps indicate that the model focuses on anatomically relevant regions of the medial clavicular epiphysis.
The proposed method, which is currently being added as part of the Skeleton-ID software (\url{https://skeleton-id.com/skeleton-id/}), is intended as a decision-support component within multi-factorial forensic workflows.
\end{abstract}



\begin{keyword}
Legal age estimation \sep Biomedical image analysis \sep Clavicle CT scans \sep 3D computer vision \sep Deep learning \sep Explainable AI
\end{keyword}

\end{frontmatter}



\section{Introduction}\label{sec1}

Assessing skeletal development is a well-established approach for estimating biological age in children and adolescents, with important applications in both clinical and forensic practice. In clinical settings, biological age assessment is routinely used in pediatric radiology and endocrinology to evaluate skeletal maturity for diagnostic and treatment-related purposes \citep{beatty2022estimating}. In forensic and medico-legal contexts, age estimation carries additional legal and social implications, particularly when determining whether an individual has reached a legally relevant age threshold, such as the age of majority \citep{refn2023prediction}.

In recent decades, large-scale migration flows and humanitarian crises have increased the number of individuals lacking reliable identity documentation, including a substantial proportion of minors \citep{migration_gov, unicef_birth}. In this context, legal age estimation (LAE) has become a critical task for governmental and judicial institutions, as age determination directly affects access to protection mechanisms, asylum procedures, and legal safeguards. Beyond migration, LAE methods are also relevant in other sensitive scenarios, such as child marriage and the investigation of crimes involving minors \citep{unicef2011child}. These applications demand methods that are not only accurate, but also reproducible, transparent, and suitable for integration into standardized forensic workflows.

Age estimation remains one of the most challenging problems in forensic anthropology \citep{ubelaker2019estimation}. Traditional approaches based on skeletal or dental development are widely used in practice, yet they often suffer from limitations including subjectivity, inter- and intra-observer variability, limited reproducibility, lack of statistical robustness and population dependence \citep{marquez2015overview, marconi2022validity}. Moreover, many methods are time-consuming and difficult to validate across heterogeneous populations. In recent years, artificial intelligence (AI) has been proposed as a promising alternative to address several of these issues by enabling objective, fast, and easily replicable estimation procedures. When trained on sufficiently large and diverse datasets, AI-based methods have the potential to improve precision and reduce observer-dependent variability \citep{mesejo2020survey}.

\begin{figure}[h!]
    \centering
        \subfloat[Position of the MCE]{
            \includegraphics[width=0.3\columnwidth]{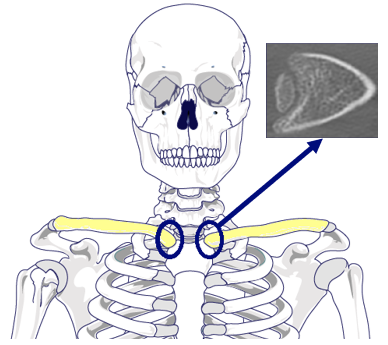}
        }
        \hfill
        \subfloat[Clavicle ossification stages]{\includegraphics[width=0.65\columnwidth]{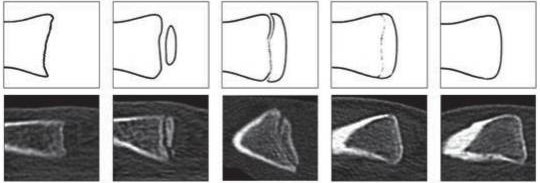}}
    \caption{Stages of clavicle ossification as proposed by \citet{kellinghaus2010forensic}. All images show the MCE, i.e., the region where the clavicle joins the sternum. Individuals from stages 1 to 5 had a mean age of 13.28, 17.81, 21.73, 29.63 and 31.77 years old for males, and of 12.70, 16.28, 21.14, 28.21 and 30.88 years old for females in the work that first applied the method \citep{kellinghaus2010forensic}. This method was later extended by dividing stages two and three into three sub-stages each. Image reproduced from \citet{schmeling2016forensic}.}
    \label{fig:clavicle_ossificatio}
\end{figure}

For LAE specifically, international forensic guidelines recommend a multi-factorial approach that combines physical examination with imaging-based assessments, including an X-ray of the hand and a panoramic radiograph of the teeth \citep{cunha2009problem, schmeling2016forensic}. Additionally, when hand skeletal development is complete, a computed tomography (CT) analysis of the clavicles, and particularly the medial clavicular epiphysis (MCE), is recommended, as the clavicle is the last bone to ossify in the human skeleton \citep{schmeling2016forensic}. When dealing with LAE using the clavicle, human experts focus on the ossification of the MCE (the part of the clavicle that is closer to the sternum), dividing it into different developmental stages \citep{kellinghaus2010forensic, kellinghaus2010enhanced} (see \autoref{fig:clavicle_ossificatio}). Different authors \citep{wesp2024radiological, qiu2024machine} report that forensic experts obtain a mean absolute error (MAE) of approximately 1.90 years for LAE using the MCE.

While AI-based LAE methods using hand radiographs or dental images have received considerable attention \citep{mesejo2020survey}, AI-driven approaches using clavicle CTs remain comparatively underexplored. To date, only a small number of studies have investigated this modality, and most focus either on limited slice-selection strategies or on end-to-end models without explicit uncertainty quantification \citep{qiu2024machine, wesp2024radiological, sun2025chronological}. More specifically, \citet{qiu2024machine} used a 2D convolutional neural network (CNN) receiving stacked slices of the previously detected MCE as input channels, while \citet{wesp2024radiological} and \citet{sun2025chronological} used 3D CNNs that received the complete MCE volume as input. Among these works, only \citet{qiu2024machine} and \citet{wesp2024radiological} focused on LAE, achieving a MAE of 1.77 and 1.65 years, respectively. Regarding \citet{sun2025chronological}, the authors did not focus on LAE, but obtained the Kellinghaus stage \citep{kellinghaus2010forensic} on a reduced dataset, achieving an accuracy of 90\%.

From an algorithmic perspective, LAE from clavicle CTs presents several open challenges. These include: (i) the automatic localization of the clavicles and the MCE in CT scans; (ii) the selection of anatomically informative slices from volumetric data, where relevant regions are sparse and cannot be reliably predefined without introducing expert bias; (iii) the design of learning architectures capable of exploiting multi-slice information under limited data regimes; and (iv) the quantification of predictive uncertainty in a manner compatible with forensic decision-making. Addressing these challenges requires integrated pipelines rather than isolated model components.

From a forensic standpoint, LAE differs fundamentally from standard age regression tasks, as the primary objective is risk-aware threshold decision-making rather than point estimation accuracy alone. Methods must be accurate and robust, but also provide uncertainty estimates that allow conservative decision-making, such as determining whether an individual can be considered at least 18 years old with a specified level of confidence. This distinction motivates approaches that explicitly link continuous age estimates to decision-aware uncertainty modeling.

In this work, we propose an explainable, multi-stage pipeline for LAE from clavicle CT scans that explicitly addresses these challenges. The methodological contributions of this work are:\begin{itemize}
    \item \textit{Automatic clavicle and MCE localization}: we introduce a feature-based connected-component method for detecting clavicles and localizing the MCE in full-body CT scans, requiring minimal manual annotation and achieving a near-perfect detection accuracy on held-out test data.
    \item \textit{Explainable multi-slice age estimation}: we propose a multi-slice convolutional learning framework in which Integrated Gradients (IG) \citep{IntegratedGradients} are explicitly incorporated into the training pipeline to guide slice selection, allowing the network itself to learn and exploit anatomically informative regions rather than using IG solely as a post-hoc visualization tool.
    \item \textit{State-of-the-art performance}: our multi-slice, multi-view CNN framework achieves superior MAE (1.55 $\pm$ 0.16 years\footnote{The 95\% confidence interval over the test MAE was estimated using 10,000 bootstrap resamples of the absolute errors.}) compared to previously published AI-based methods on the same task \citep{wesp2024radiological, qiu2024machine, sun2025chronological} (1.65-1.77 years reported, 1.75-2.30 years in our dataset for simplified implementations). These differences are statistically significant in favor of our method, demonstrating that careful anatomical guidance and slice selection substantially improve generalization. Performance is also superior to forensic experts that obtain, approximately, a MAE of 1.90 years \citep{wesp2024radiological, qiu2024machine}.
    \item \textit{Decision-aware uncertainty quantification}: we incorporate conformal prediction to obtain calibrated prediction intervals, enabling the application of the minimum-age concept and facilitating integration into multi-factorial forensic workflows. This method is currently being integrated into the commercial Skeleton-ID software \citep{skeletonid}, enabling application in real forensic cases.
\end{itemize}


\section{Proposed method}

\subsection{System overview}

\begin{figure*}[h!]
    \centering
    \includegraphics[width=\linewidth]{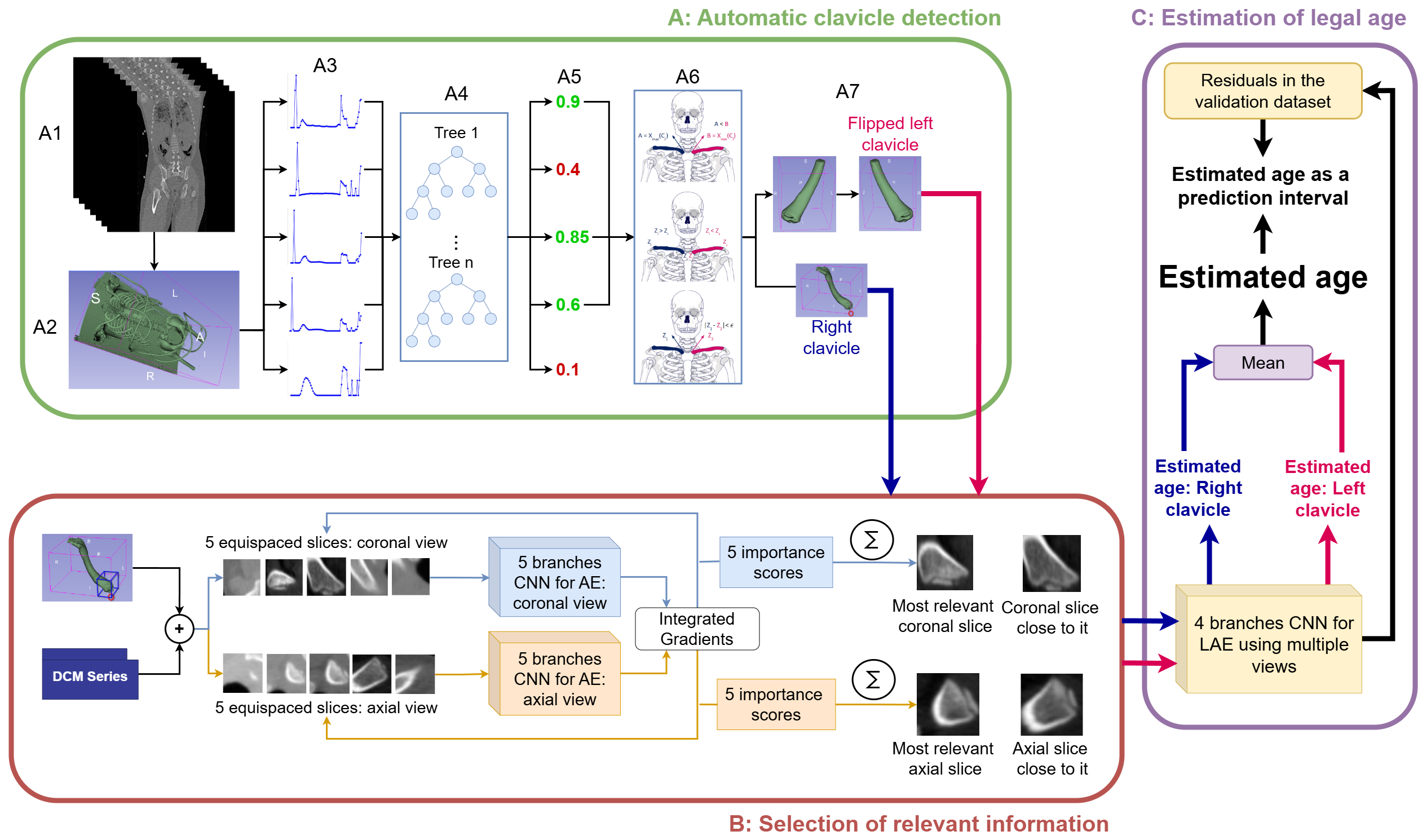}
    \caption{Overview of the proposed age estimation method. From a full-body CT (A1), the skeleton is automatically segmented by thresholding at 300 Hounsfield Units (A2). All connected components in A2 are extracted, and shape and geometrical features are computed to form a feature vector (A3), which is input to a Random Forest (A4) that estimates the probability of each component being a clavicle (A5). The three components with the highest probabilities are then filtered using anatomical constraints (A6) to ensure a maximum of two clavicles per individual (A7). Each identified clavicle is processed in part B, where the MCE is localized in the original CT using the segmented clavicle. Five equispaced coronal slices are then passed through a five-branch CNN trained for age estimation. Integrated Gradients \citep{IntegratedGradients} assigns an importance score to each branch, which is used to compute a weighted average of slice positions to identify the most relevant slice. The same procedure is applied to the axial view, while the sagittal view is omitted as it did not improve performance (see \autoref{results_sec}). The most relevant slices and neighboring slices are input to the LAE model in step C. This process is performed for both clavicles, and the final age estimate is the average of the two. Using the conformal split method \citep{lei2018distribution}, a prediction interval for the estimated age is also obtained.}
    \label{fig:the_method}
\end{figure*}

The proposed LAE framework consists of a multi-stage pipeline designed to process full-body CT scans and produce age estimates with calibrated uncertainty. As illustrated in \autoref{fig:the_method}, the pipeline comprises the following components:\begin{itemize}
    \item \textit{Automatic clavicle detection}: we use connected-component analysis combined with feature-based classification and anatomy-guided constraints to detect both the right and left clavicles.
    \item \textit{Selection of relevant information}: the MCE is localized inside the detected clavicle via an anatomically informed bounding box. The yielded volume is provided to an explainable slice selection method guided by IG \citep{IntegratedGradients}. For this purpose, IG is integrated into the data selection stage rather than used solely for post-hoc interpretation.
    \item \textit{Estimation of legal age}: first, the selected CT slices are forwarded through a multi-slice CNN for age estimation. When both clavicles are successfully detected, their age estimates are integrated using the mean. Moreover, conformal prediction is used to quantify uncertainty and to support minimum-age decisions.
\end{itemize}

The output of the pipeline is a point estimate of chronological age together with a calibrated prediction interval, which can be directly used to support minimum-age (e.g., $\geq$18 years) decisions. Moreover, the pipeline is designed to decouple anatomical localization from age regression and improve robustness under limited training data.

\subsection{Automatic detection of the MCE}\label{sec:clav_detection}

Since our dataset consisted of full-body CTs, the first step in developing a LAE method using the MCE was to detect this region. Moreover, this detection method should not only be applicable to full-body CTs, but also to thoracic CTs, which are much more common in LAE using the clavicle. In order to do this, we leveraged two key observations: (i) the human skeleton can be represented as a graph of connected components (individual bones or bone groups), including the left and right clavicles; and (ii) clavicles have distinctive shape and geometrical properties that distinguish them from other bones.

Skeleton segmentation was performed by thresholding the CT at 300 Hounsfield Units. This threshold is a widely accepted standard in musculoskeletal imaging to suppress soft tissues while preserving cortical and trabecular bone structures \citep{ma2022x}. We retained all voxels above the threshold as bone and transformed the matrix into a mesh using the marching cubes algorithm \citep{marchingcubes}. Mesh reconstruction was required to compute rotation-invariant shape descriptors and surface-based geometric features, which are not directly available in the voxel domain. Connected components were extracted from the mesh. 

For each connected component, we extracted a feature vector encoding shape and geometry. Shape was represented using shape distributions, excluding the computationally expensive \(D4\) distribution \citep{osada2002shape}. Geometrical features included area, volume, normalized shape index, area-to-volume ratio, length-to-width ratio, and sphericity. Note that since these features are directly obtained from the segmented clavicle mesh, which preserves the original clavicle size and shape, they are applicable independently of the CT of origin, whether it is a full-body or a thoracic CT.

Clavicle detection was formulated as a supervised binary classification problem. For each connected component, the described feature vector was computed and used for classification. Several classifiers were compared using 5-fold cross-validation on a labeled subset of the training data (clavicle vs. non-clavicle), and a Random Forest (RF) \citep{ho95} achieved the best performance. The final RF consisted of 50 trees, with $\sqrt{m}$ features considered at each split and entropy used as the split criterion. Hyperparameters were selected based on cross-validation performance. The final model was trained on the full training set and evaluated on a held-out test set.

\begin{figure}[h!]
    \centering
        \subfloat[There are, potentially, a left and a right clavicle.]{
            \includegraphics[width=0.45\columnwidth]{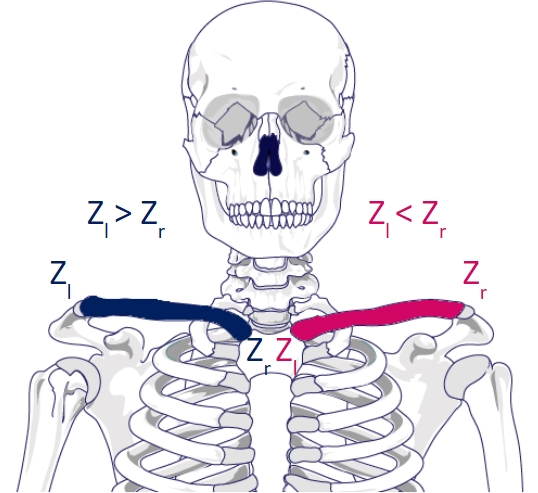}
        } \hfill
        \subfloat[Verify that each clavicle is at the corresponding side.]{
            \includegraphics[width=0.45\columnwidth]{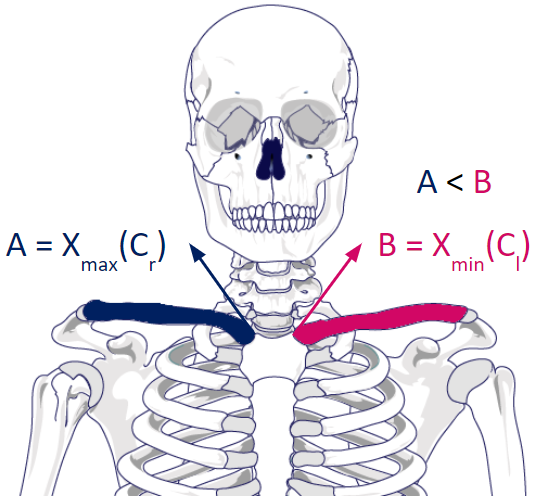}
        } \\
        \subfloat[Clavicles are close to each other.]{
            \includegraphics[width=0.45\columnwidth]{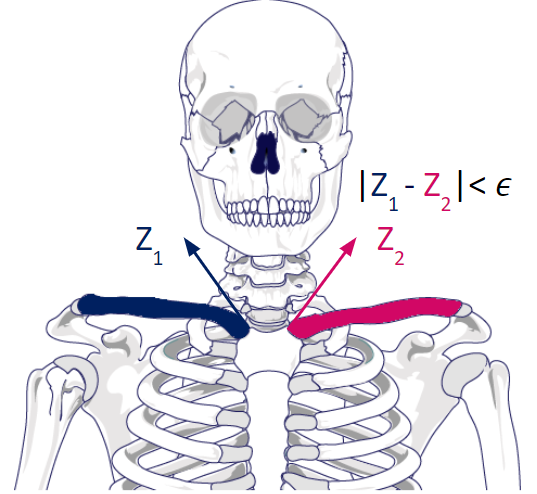}
        } \hfill
        \subfloat[Coordinate system.]{
            \includegraphics[width=0.45\columnwidth]{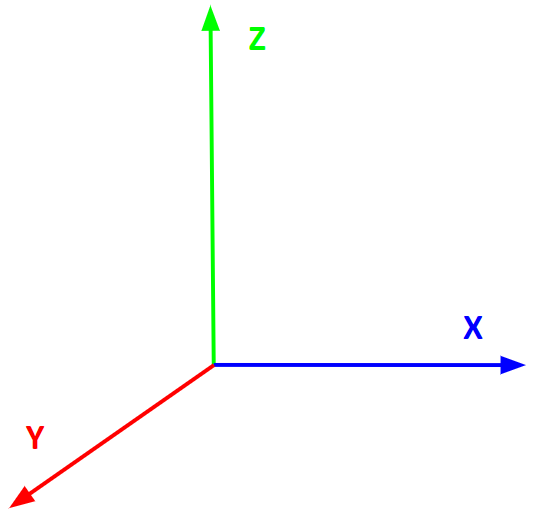}
        }
    \caption{Anatomical constraints for clavicle detection \citep{netter2014atlas}. Right clavicle is shown in blue. (a) For each clavicle, $Z_l$ is the $Z$ value at the leftmost ($X_{min}$) point, and $Z_r$ for the rightmost one ($X_{max})$. This can be used to detect clavicle laterality (left or right clavicle, denoted as $C_l$ and $C_r$ in (b). (b) Once laterality is defined, the leftmost point of $C_r$ ($A$) must be at the right ($<$) of the rightmost point of $C_l$ ($B$). (c) let $Z_1$ be the height of the innermost (leftmost) point of $C_r$, and $Z_2$ the same for $C_l$. $|Z_1 - Z_2|$ must be below a fixed distance $\epsilon$, with $\epsilon=1$cm in our case.}
    \label{fig:props}
\end{figure}

Once trained, the RF estimated the probability of each component being a clavicle. To handle occasional misses, the three highest-probability components (\(X_1, X_2, X_3\) with \(P_1 \ge P_2 \ge P_3\)) were considered. If \(P_2 \ge 0.5\) and \(P_3 < 0.5\), \(X_1\) and \(X_2\) were chosen. Otherwise, pairwise compatibility (e.g. if $X_1$ is the right clavicle then $X_2$ or $X_3$ can be the left one) was checked using anatomical constraints (see \autoref{fig:props}) in an attempt to inject domain knowledge into the detection process for \(X_1\) and \(X_2\), then \(X_1\) and \(X_3\). The first compatible pair was selected; if none satisfied all constraints, only \(X_1\) was chosen if \(P_1 > 0.5\). The threshold of 0.5 was chosen to reflect a conservative binary decision boundary. This procedure handles cases of merged or partially obscured clavicles, which are uncommon but can occur, especially when working with deceased individuals.

\begin{figure}[h!]
    \centering
        \subfloat[Innermost vertex of clavicle at \((X_{\max}, Y_{\max}, Z_{\min})\).]{
            \includegraphics[width=0.45\columnwidth]{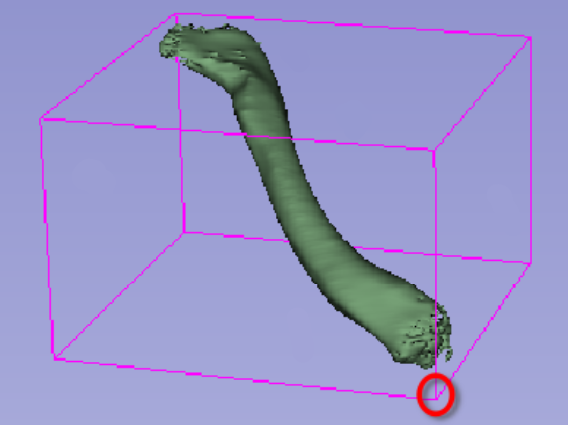}
        } \hfill
        \subfloat[Sub-bounding box around the MCE.]{
            \includegraphics[width=0.45\columnwidth]{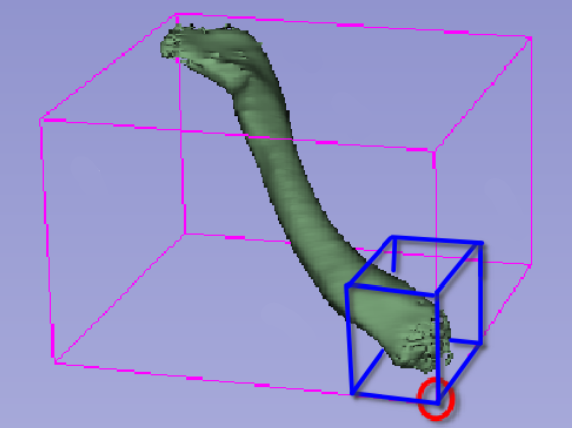}
        }
    \caption{Bounding box localization around the MCE. Because of the clavicle's anatomical positioning, it is known that the MCE is closest to the vertex of the bounding box at \((X_{\max}, Y_{\max}, Z_{\min})\), using the coordinate system shown in \autoref{fig:props}.}
    \label{fig:MCE}
\end{figure}

After detection, bounding boxes were extracted around each clavicle. The medial epiphysis was then localized within each box: for the right clavicle, the box vertex at \((X_{\max}, Y_{\max}, Z_{\min})\) was expanded by 2.5 cm along \(-X\), \(-Y\), and \(+Z\), since this is enough to include all relevant information \citep{bernat2014anatomy}. The left clavicle was processed similarly after horizontal flipping around the Z-axis. All localized MCE regions were resampled to an isotropic voxel spacing of $0.5 \times 0.5 \times 0.5$ mm using linear interpolation, in order to eliminate inter-subject variability in spatial resolution. The original image orientation, origin, and coordinate system were preserved, and no spatial transformation beyond resampling was applied. This preprocessing step eliminates bias caused by varying in-plane resolutions and ensures that the morphological features of the MCE are represented with equal fidelity in the axial, coronal, and sagittal planes. This facilitates standardized feature extraction for age estimation. For a graphical example of this process, see \autoref{fig:MCE}.

Following isotropic resampling, the MCE volumes were processed as 3D arrays from which 2D slices were extracted in the axial and coronal planes. These slices correspond to exact planar cross-sections of the standardized 3D volume and were used as inputs to the trained CNNs described in the following steps. Since no intensity projection or dimensionality reduction was applied, each slice represents a single physical plane within the resampled volume. Note that, since we cropped a region of $2.5 \times 2.5 \times 2.5$ cm, and resampled to a spacing of $0.5 \times 0.5 \times 0.5$ mm, we obtained 50 2D images per view. 

\subsection{Integrated Gradients–guided slice selection}\label{mostImportantSliceObtention}

Age estimation from clavicle CTs relies on a limited subset of slices that capture the degree of epiphyseal ossification of the MCE \citep{kellinghaus2010forensic,kellinghaus2010enhanced}. Rather than selecting slices heuristically or processing the entire volume, we propose an IG–guided slice selection strategy \citep{IntegratedGradients} that is explicitly integrated into the training pipeline.

\subsubsection{Stage 1: IG-based estimation of the most informative slice.} 
In a first stage, a multi-branch CNN is trained for each anatomical view using a fixed number of input slices. Specifically, five equispaced slices are extracted from each resampled MCE volume. Each slice is processed by an independent branch of a common CNN, where each branch is a ResNet50 pre-trained on ImageNet \citep{resnet,ImageNet}. The outputs of the five branches are concatenated and passed to a fusion network composed of fully connected layers that regress chronological age.

Once trained, this model is not used for age estimation, but rather as a slice-importance estimator. For each subject, IG is computed with respect to the input of the fusion network, i.e., the concatenated feature vector formed by the outputs of the five branches. This yields an attribution score for each feature dimension, reflecting its contribution to the predicted age.

Let \(S_n\) denote the aggregated IG attribution score of branch \(n\), obtained by summing IG values over all feature dimensions associated with that branch. Since each branch processes a known slice position \(P_n\) within the input volume, the estimated position \(c\) of the most informative slice is computed as a weighted average:
$$
c = \frac{\sum_{n=0}^{4} S_n \cdot P_n}{\sum_{n=0}^{4} S_n}.
$$
This procedure yields, for each subject and each view, a continuous estimate of the slice position that contributes most strongly to age prediction.

\subsubsection{Stage 2: Construction of IG-centered training samples.}
In the second stage, the estimated central slice \(c\) is used to construct the training inputs for the final age estimation model. Rather than using a single slice, a set of \(m\) slices is selected to capture complementary anatomical context. These slices are chosen to be equispaced within the interval \([c-a, c+b]\), where \(a\) and \(b\) are non-negative hyperparameters controlling the inclusion of slices before and after the central position.

This strategy reflects expert practice in forensic age estimation, where a diagnostically relevant slice is first identified and then examined together with neighboring slices. Based on validation experiments, we set \(m=2\), \(a=0\), and \(b=10\) for both coronal and axial views. Sagittal slices were not used due to their lower relevance in age estimation, as shown in \autoref{results_sec}.

IG is used here not as a post-hoc interpretability tool, but as a data-driven mechanism for constructing training samples, thereby aligning the model’s inputs with anatomically relevant regions of the clavicle.

\subsection{Legal age estimation using IG-guided multi-slice CNNs}\label{AEApproaches}

After IG-guided slice selection, LAE is performed using a multi-slice, multi-branch deep learning model. For each anatomical view, the selected slices are processed by a dedicated CNN in which each slice is handled by an independent branch. Each branch consists of a ResNet50 \citep{resnet} backbone pre-trained on ImageNet \citep{ImageNet}, used as a feature extractor without weight sharing between branches. Note that this architecture is similar to the one used for IG-guided slice selection, but is used with a different purpose: LAE over filtered informative slices.

The output feature vectors of all branches are concatenated to form a joint representation, which is passed to a fusion network composed of fully connected layers with ReLU activations and dropout regularization \citep{dropout}. The number of consecutive blocks of this type is a hyperparameter that we named classifier depth. The fusion network regresses a single continuous age estimate. This design allows the network to first learn discriminative representations for individual slices and then model interactions between slices at a higher semantic level.

This architecture was selected over alternatives based on stacked feature maps, as it provides greater flexibility in modeling slice-specific information and enables direct estimation of slice importance during the IG-guided selection stage.

\begin{figure}[h!]
    \centering
    \includegraphics[width=\linewidth]{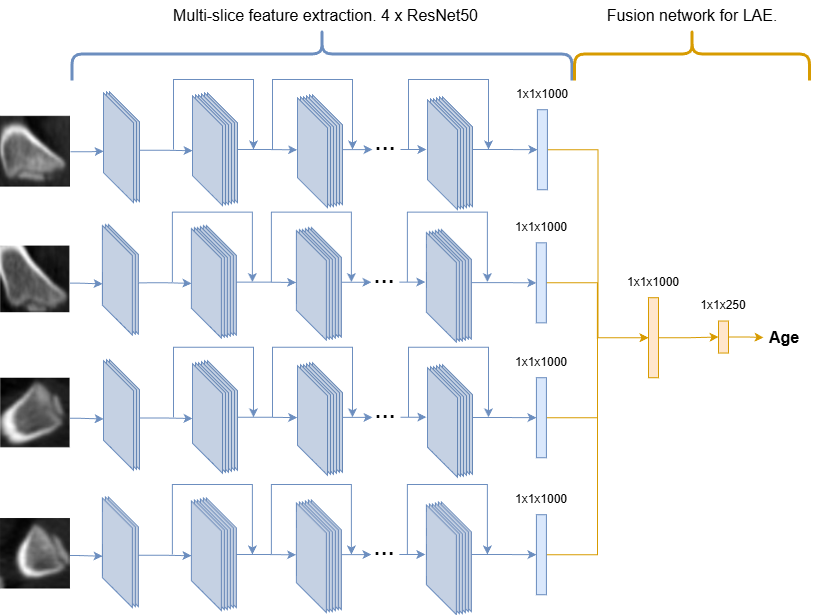}
    \caption{Simplified representation of the proposed architecture used for LAE. It consists of 4 branches that process different slices (2 coronal and 2 axial) using ResNet50 \citep{resnet}. The output of all branches, consisting of a 1,000-dimensional feature vector is passed to a fully-connected network that we named the fusion network. By doing this, we obtain an end-to-end model for LAE using multi-slice learning.}
    \label{fig:multislice}
\end{figure}

As stated in the previous subsection, the best model uses two axial and two coronal slices, so it has four branches, where each of them outputs 1,000 features. As for the fusion part of the network, it has a depth of 2, containing: a fully-connected layer receiving 4,000 features and outputting 1,000 features (dropout rate of 0.25), a ReLU activation, a fully-connected layer outputting 250 features (dropout rate of 0.25), a ReLU activation, and a fully-connected layer that estimates age (dropout rate of 0.5). A schematic overview of the multi-slice architecture is shown in \autoref{fig:multislice}.

\subsubsection{Combining age estimates from both clavicles}\label{CombinationMethods}

For most individuals, both left and right clavicles were available, enabling the combination of age estimates from the two bones to potentially improve accuracy. Although we experimented with different alternatives (as will be explained in \autoref{sec:merge_method_comparison}), we concluded that the best option is to estimate age using each clavicle separately, then compute the mean of both estimations. When only one clavicle is available (because the method failed to detect the other one), the estimated age is provided as the final one. 

\subsection{Prediction intervals and the minimum age concept}\label{intervals_method}

European protocols recommend using the minimum age concept when estimating age \citep{schmeling2016forensic}. Specifically, when multiple characteristics or bones are assessed, the highest estimated minimum age should be used as the final legal-age decision. Therefore, it is not sufficient to provide only a point estimate of age, and a calibrated prediction interval is also required to support minimum-age determinations.

To obtain such intervals, we employed the \emph{conformal split} method \citep{lei2018distribution}. Let $\hat{y}_i$ denote the predicted age for individual $i$ in the validation set, and $y_i$ the true chronological age. We first compute the absolute residuals in validation:

$$
r_i = |\hat{y}_i - y_i|.
$$

For a desired coverage level $\beta = 1 - \alpha$ (e.g. 0.9 or 0.95), we select the $100\beta$\% smallest residuals from the validation set and compute their maximum value, denoted $d_\beta$. This value represents the radius of the prediction interval that guarantees coverage $\beta$ under the conformal framework.

For a new test prediction $\hat{y}$, the conformal prediction interval is then defined as:

$$
[\hat{y} - d_\beta, \hat{y} + d_\beta].
$$

This interval provides a calibrated estimate of uncertainty and can be directly interpreted in the context of minimum-age decisions: the lower bound of the interval can be considered a conservative estimate of the individual’s minimum age, ensuring compliance with legal forensic guidelines.

\subsection{Training details}\label{TraningMethods}

All age estimation models were trained using the Adam optimizer \citep{adam} with $\beta_1=0.9$ and $\beta_2=0.999$ and the mean squared error as the loss function. Training was performed in two stages. 

First, the fusion part of the multi-branch CNN, i.e., the sequence of fully connected layers that receives the output of all branches and estimates age, was trained for one epoch with a learning rate of $\eta_1$ (this being a hyperparameter), while keeping all branch weights frozen. This allows the fusion layers to adapt to the feature representations produced by the pre-trained branches.

Afterwards, we fine-tuned the model. For this purpose we used a learning rate of $\eta_2 = \eta_1/100$ to train: (i) the last convolutional group of each branch, which in ResNet50 \citep{resnet} corresponds to the final three residual blocks; (ii) the terminal fully-connected part of each branch; and (iii) the fusion part of the multi-branch CNN.

In both training stages, we used the \emph{1cycle policy} \citep{cyclical_lr, one_cycle}. In this case, each layer $i$ with a learning rate $\eta_i$ started at a fraction of it $\eta_i/25$, increased to the base value $\eta_i$, and then decreased back to $\eta_i/1000$ over a single cycle.

These and other hyperparameters, such as batch size, number of epochs, and early stopping criteria were selected based on validation performance. Our best experiment used a batch size of 4 and was trained with $\eta_1 = 0.001$ for 34 epochs (1 of which was used for the first stage), halting due to early stopping. Data augmentation, described in \autoref{DataAugmentation}, was applied during training to improve robustness.

\subsection{Data augmentation}\label{DataAugmentation}

To increase the effective size of the training dataset and improve robustness, we developed a dedicated slice-level data augmentation strategy tailored for multi-slice MCE inputs. Each training sample consists of $m$ equispaced slices extracted from a 3D bounding box around the MCE, centered on the most informative slice $c$ identified via the IG-guided procedure (\autoref{mostImportantSliceObtention}).

In the standard (non-augmented) setting, the $m$ slices are selected symmetrically around $c$. For instance, for $m=5$, the slices might be $\{c-10, c-5, c, c+5, c+10\}$. To augment the data, we shift the central slice slightly in $-i$ or $+i$, treating positions $c-i$ and $c+i$ as alternative centers, and then select $m$ equispaced slices around these positions. Using the previous example, if the central slice is shifted to $c-2$, the resulting set would be $\{c-12, c-7, c-2, c+3, c+8\}$. This procedure is repeated for multiple offsets (-2, +2, -4 and +4 in the experiment with best results), yielding multiple, slightly different sets of slices per individual. This augmentation preserves the anatomical context while introducing variability that helps the model generalize.

This approach leverages the inherent redundancy of adjacent MCE slices, mimicking the process of a human expert who inspects neighboring slices around a central reference slice when estimating age. As a result, each training individual contributes multiple effective training samples, improving model robustness without artificially distorting the anatomical features of the clavicle.



\section{Experiments}\label{results_sec}

\subsection{Used dataset}
This study uses a publicly available dataset of 1,158 full-body post-mortem CT scans obtained from the New Mexico Decedent Image Database (NMDID) \citep{edgar2020new}. The dataset includes individuals aged between 14.0 and 25.99 years (mean age: 21.31 years), covering the age range relevant for LAE based on clavicular development.

Following forensic recommendations for assessing age using the MCE, we selected CT acquisitions with thin slice thickness (1 mm) and sufficient spatial resolution. Specifically, we used the torso CT series labeled THIN\_ST\_TORSO, THIN\_ST\_UEXT-TO, or TH\_ST\_TORSO in the database. All scans were acquired with comparable imaging protocols, including a tube voltage of 120 kVp and tube current of approximately 300 mAS. All selected CT series were reconstructed with a slice thickness of 1.0 mm and an axial slice spacing of 0.5 mm, resulting in overlapping slices along the z-axis. In-plane matrices were 512×512, with pixel spacing ranging from 0.8 to 1.2 mm depending on the field of view. These acquisition parameters are consistent with forensic recommendations for evaluating the MCE.

The dataset was split into training (742 scans), validation (184 scans), and test (232 scans) sets using stratified sampling to preserve the age distribution across splits. Splitting at the individual level prevented data leakage (e.g., avoiding the left clavicle in training and the right clavicle of the same individual in testing).

Since LAE is performed at the level of the clavicle rather than the full scan, each CT was processed using the detection method proposed in \autoref{sec:clav_detection} to extract the left and right clavicles when present. A manually curated subset of the training scans containing 160 individuals was used to train (128) and evaluate (32) this method. For each individual the connected components (3D bone meshes) were shown to a human annotator who identified the clavicles. The resulting labeled dataset comprised 317 clavicles and 2,075 non-clavicle bones for model training and validation (we used 5-fold cross-validation), and 54 clavicles and 359 non-clavicle bones for testing.

After running the clavicle detection process over the complete dataset, we obtained 1,447 clavicles for training, 353 for validation, and 457 for testing.

\begin{figure}[h!]
    \centering
    \includegraphics[width=\columnwidth]{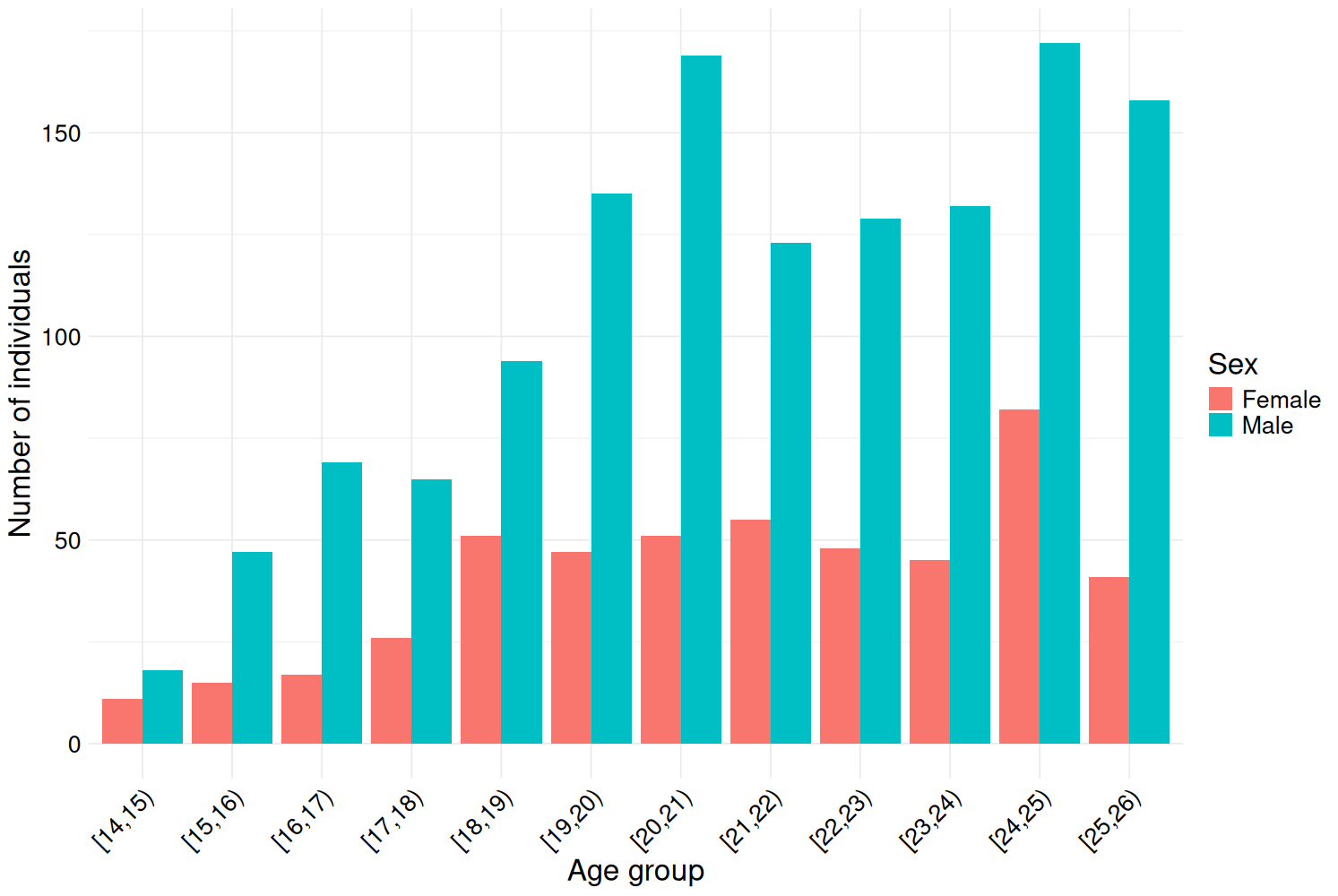}
    \caption{Age and sex distribution of the dataset employed for training and validating our LAE method. Both chronological age and sex were specified in the metadata of the downloaded cases.}
    \label{fig:distr}
\end{figure}

Two relevant imbalances can be observed in the age and sex distribution shown in \autoref{fig:distr}: (i) a limited number of individuals below the 18-year threshold, and (ii) a predominance of male subjects (73\%). These characteristics reflect real-world forensic scenarios in which clavicle-based age estimation is typically applied to older adolescents and young adults, and where male subjects are overrepresented. That said, to mitigate the impact of these imbalances during training, we tested the performance of applying targeted data augmentation to underrepresented age groups, while also reporting stratified performance metrics to assess potential biases. Augmentation techniques are explained in \autoref{DataAugmentation}.

\subsection{Clavicle detection: ablation study}

To assess the contribution of the different feature groups introduced in \autoref{sec:clav_detection}, we first defined a simple distance-based baseline that does not rely on supervised learning. Using 10 manually identified clavicles from the training set, we computed an average clavicle representation by averaging their feature vectors. For each individual, the distance between every segmented bone and this average clavicle was computed (using the Wasserstein distance for shape features and the Euclidean distance for the remaining ones). Based on the anatomical prior that at most two clavicles are present per individual, the two closest components were labeled as clavicles. Starting from this baseline, we progressively introduced three components of the proposed method in order to quantify their individual contribution: (i) replacing the distance-based rule with a supervised machine learning classifier; (ii) selecting the two connected components with the highest predicted clavicle probability instead of relying on hard class labels; and (iii) incorporating lightweight anatomical constraints to remove anatomically implausible configurations. The complete method achieved the best performance across all metrics in cross-validation, with an accuracy of 99.80\%, sensitivity of 98.86\%, and specificity of 99.94\%. Results of the ablation study are shown in \autoref{tab:selection}. When applied to the independent test set (413 bones, including 54 clavicles), the proposed method achieved 100\% accuracy, with no misclassified clavicles. We evaluated several classifiers, including Decision Trees \citep{quinlan1986induction}, RF \citep{ho95}, Support Vector Machines \citep{svmorig} and Logistic Regression \citep{LogisticRegression}. RF achieved the highest balanced accuracy and was therefore selected for all subsequent experiments.

\begin{table}[!h]
\footnotesize
\centering
\caption{Results of the ablation study conducted during the development of the clavicle detection method. The positive class is the connected component (bone) being a clavicle. Best alternative in bold, corresponding to the highest average validation values in cross validation. Every row of the table represents an addition over the previous row.}
\label{tab:selection}
\begin{tabular*}{\linewidth}{@{\extracolsep{\fill}} cccc }
\toprule
 Method                        & Accuracy & Sensitivity & Specificity \\ \midrule
Distance-based shape features             & 93.33\%  & 76.68\%     & 95.91\%     \\
Add geometric features             & 94.29\%  & 80.27\%     & 96.46\%     \\
Add RF classifier            & 98.84\%  & 93.16\%     & 99.71\%     \\
Only take 2 of highest probability & 99.04\%  & 97.72\%     & 99.24\%     \\
\textbf{Add anatomical constraints} & \textbf{99.85\%}  & \textbf{99.24\%}     & \textbf{99.94\%}     \\ \bottomrule
\end{tabular*}
\end{table}

\subsection{Multi-slice-based age estimation: ablation study}\label{sec:ae_res}

To quantify the contribution of the different components of the proposed LAE framework, we conducted a structured ablation study whose results are summarized in \autoref{tab:experimentss}. Unless otherwise stated, all models shared the same training protocol and used ResNet50 \citep{resnet} as the backbone of each CNN branch. After selecting the best-performing configuration in \autoref{tab:experimentss} (validation MAE of 1.43 years), we evaluated alternative CNN backbones. Using ResNet18 \citep{resnet}, ConvNeXt-T \citep{ConvNeXt}, ResNeXt50 \citep{resneXt}, VGG16 \citep{vgg}, and EfficientNet-V2-S \citep{efficientnetv2} resulted in validation MAEs of 1.49, 1.45, 1.47, 1.67, and 1.49 years, respectively. Although some architectures achieved lower training error, none improved validation performance, indicating that ResNet50 provides a favorable trade-off between model capacity and generalization for this task. Additional ablations showed that several commonly used refinements did not improve performance in this setting. Specifically, employing a differential learning rate, incorporating sex as an additional input, freezing batch normalization layers, or training independent single-slice models all resulted in higher validation MAEs (1.46–1.49 years), confirming that the proposed multi-slice, end-to-end configuration is preferable.

\begin{table*}[h!]
\footnotesize
\centering
\caption{Ablation study of the multi-branch LAE CNN evaluated on the validation set. Groups of experiments correspond to controlled modifications of a single component (view, slice selection, number of slices, data augmentation, or classifier depth). The best-performing configuration is highlighted in bold.}
\label{tab:experimentss}
\definecolor{tablegray}{gray}{0.95} 

\begin{tabularx}{\linewidth}{l c c c c c c Y} 
\toprule
\textbf{View} & \begin{tabular}[c]{@{}c@{}}\textbf{Num.} \textbf{slices}\end{tabular} & \begin{tabular}[c]{@{}c@{}}\textbf{Data} \textbf{augmentation}\end{tabular} & \begin{tabular}[c]{@{}c@{}}\textbf{Classifier} \textbf{depth}\end{tabular} & \begin{tabular}[c]{@{}c@{}}\textbf{Slice} \textbf{selection}\end{tabular} & \begin{tabular}[c]{@{}c@{}}\textbf{MAE} \textbf{train}\end{tabular} & \begin{tabular}[c]{@{}c@{}}\textbf{MAE} \textbf{val}\end{tabular} & \textbf{Conclusions} \\ 
\midrule

Axial & 5 & No & 1 & No & 0.41 & 1.78 & \multirow{2}{=}{\centering Slice selection improves results} \\
Axial & 5 & No & 1 & Yes & 0.52 & 1.69 & \\ \midrule

Coronal & 5 & No & 1 & Yes & 0.54 & 1.66 & \multirow{2}{=}{\centering Coronal $>$ axial $>$ sagittal} \\
Sagittal & 5 & No & 1 & Yes & 0.49 & 1.73 & \\ \midrule

Coronal & 5 & Yes & 1 & Yes & 0.36 & 1.49 & \multirow{2}{=}{\centering Data augmentation improves results} \\
Coronal & 5 & Balance & 1 & Yes & 0.42 & 1.65 & \\ \midrule

All & 3 & Yes & 1 & Yes & 0.46 & 1.49 & \multirow{2}{=}{\centering Ignore the sagittal view} \\
Axial + coronal & 2 & Yes & 1 & Yes & 0.46 & 1.48 & \\ \midrule

Axial + coronal & 4 & Yes & 1 & Yes & 0.45 & 1.45 & \multirow{2}{=}{\centering Use 2 slices per view} \\
Axial + coronal & 6 & Yes & 3 & Yes & 0.55 & 1.47 & \\ \midrule

\textbf{Axial + coronal} & \textbf{4} & \textbf{Yes} & \textbf{2} & \textbf{Yes} & \textbf{0.61} & \textbf{1.43} & \multirow{3}{=}{\centering Use a classifier depth of 2} \\
Axial + coronal & 4 & Yes & 3 & Yes & 0.48 & 1.50 & \\
Axial + coronal & 4 & Yes & 4 & Yes & 0.50 & 1.53 & \\

\bottomrule
\end{tabularx}
\end{table*}

Based on these experiments, we selected a final configuration consisting of four branches: two coronal and two axial slices centered around the most relevant slices identified by the proposed slice selection strategy. The model was trained using our data augmentation strategy, and its fusion part uses a classifier composed of two fully connected blocks before the final regression layer, each consisting of a linear layer with dropout \citep{dropout}, and a ReLU activation.

The slice selection mechanism showed high anatomical consistency across individuals, indicating that the model reliably focuses on a stable and anatomically meaningful region of the MCE. In the axial view, the same slice was selected in approximately 95\% of cases across training, validation, and test sets, while coronal selections clustered within a narrow anatomical range around the MCE, in which the most selected slice was picked in ~43\% of cases across training, validation, and test.

Although age-range balancing via data augmentation improved performance for the youngest and oldest individuals, it degraded accuracy in the critical 17–25 age range, increasing overall validation MAE to 1.65 years. Given that individuals under 18 are typically identified as minors prior to clavicle assessment \citep{schmeling2016forensic}, and that precision is less critical at older ages, we retained the unbalanced training strategy. Age-stratified errors in the test set are reported in \autoref{sec:test_res}.

\subsection{Fusing the age estimates of both clavicles}\label{sec:merge_method_comparison}

Once the best-performing single-clavicle model was selected (validation MAE of 1.43 years), we evaluated whether combining age estimates from both clavicles could further improve performance. Three alternative fusion strategies were experimentally compared.

First, we trained a joint model that processes both clavicles simultaneously by doubling the number of CNN branches. This approach reduced the training MAE to 0.50 years (compared to 0.61 years for the single-clavicle model), while increasing the validation MAE to 1.65 years. This divergence between training and validation performance indicates overfitting, which can be attributed to the increased input dimensionality combined with an effective reduction in the number of independent training samples.

Second, we evaluated a simple averaging strategy, where the left and right clavicles are processed independently and the final age estimate is computed as the mean of both predictions. When restricting the analysis to individuals for whom both clavicles were available, this approach reduced the training and validation MAEs to 0.49 and 1.27 years, respectively. When also including individuals with only one available clavicle (15 in validation and 37 in training), the validation MAE increased slightly to 1.34 years, while the training MAE remained unchanged at 0.49 years.

Finally, we explored learning a combination of both clavicle estimates using a separate machine learning model. Because this model was trained using leave-one-out cross-validation on the validation set, only validation MAEs are reported for this strategy. Among the evaluated models, a simple linear regression achieved the best performance, with a validation MAE of 1.36 years, both with and without including sex as an additional feature.

Given its superior empirical performance and simplicity, averaging the age estimates of both clavicles was selected as the final fusion strategy.

\subsection{Results obtained by alternative approaches}
Before converging on the proposed multi-slice framework, we explored more direct modeling strategies that have been commonly used in medical image analysis. These alternatives were ultimately discarded due to inferior generalization performance. After detecting the MCE or extracting the full clavicle as described in \autoref{sec:clav_detection}, two main classes of approaches were evaluated.

The first approach consisted of training 3D CNNs directly on CT volumes. We evaluated several 3D ResNet architectures (ResNet10, 18, 34, 50, and 101 \citep{resnet, monai}) pre-trained on MedicalNet \citep{medicalnet}. Standard volumetric data augmentation techniques were applied, including random scaling, rotation, contrast variation, and Gaussian noise injection. Despite these measures, all 3D CNN configurations exhibited pronounced overfitting: training error decreased steadily, while validation MAE plateaued between 1.8 and 1.9 years. The best performing configuration fine-tuned a 3D ResNet50 on the MCE region, achieving a validation MAE of 1.75 years. When extending the input to include the entire clavicle, performance degraded substantially, yielding a validation MAE of 2.40 years.

The second alternative treated multiple 2D CT slices as channels of a single multi-channel image, resulting in an input with $m$ channels, where $m$ denotes the number of slices. To leverage ImageNet-pretrained 2D CNNs \citep{ImageNet}, two strategies were investigated: (i) reducing the number of input channels from $m$ to three via an initial downsampling layer, or (ii) modifying the first convolutional layer to accept $m$ channels, initializing the additional channels by replicating the pretrained weights. The latter strategy yielded the best performance among these variants, with a validation MAE of 2.30 years. Nevertheless, this approach remained substantially inferior to the proposed method, indicating that stacking slices as channels fails to adequately capture the spatial and anatomical relationships across views.

Overall, both alternative paradigms performed markedly worse than the proposed multi-slice, multi-branch approach, which achieved a validation MAE of 1.34 years. These results highlight the advantage of explicitly modeling selected anatomical views while maintaining a moderate model complexity, enabling improved generalization in a setting with limited data.

\subsection{Results in test}\label{sec:test_res}

Using the final configuration selected in \autoref{sec:ae_res}, we evaluated the complete pipeline on the independent test set. Age was first estimated separately for each clavicle using the selected multi-slice model, and the final age estimate for each individual was obtained by averaging both predictions when available. This approach yielded a MAE of 1.55 $\pm$ 0.16 years and an $R^2$ value of 0.57 on the test set. Although the $R^2$ value is moderate, the model achieves a low MAE, comparable to expert performance and prior studies. The relatively limited $R^2$ likely reflects intrinsic biological variability and the restricted age range of the cohort.

\begin{figure}[h!]
    \centering
    \includegraphics[width=\columnwidth]{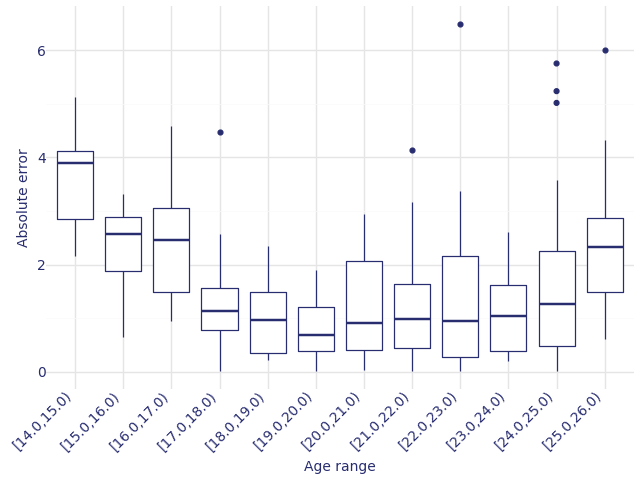}
    \caption{Absolute error of our model per age range in test.}
    \label{fig:test_box}
\end{figure}

Performance was consistent across sexes, with a MAE of 1.56 years for males and 1.52 years for females. Age-stratified errors are shown in \autoref{fig:test_box}. As expected, higher errors were observed at the extremes of the age range, particularly for individuals younger than 17 and older than 25, while substantially lower errors were obtained in the clinically and forensically most relevant interval between 17 and 25 years. The regression plot further shows a tendency to overestimate age in younger individuals and underestimate age in older individuals.

As described in \autoref{intervals_method}, prediction intervals were computed using conformal prediction in order to support the application of the minimum age concept recommended by European protocols \citep{schmeling2016forensic}. As expected, increasing coverage resulted in a trade-off with adult classification performance (see \autoref{tab:pred_intervals}). The choice of coverage level should therefore be determined by the forensic expert based on the desired balance between sensitivity for minors and precision for adults. Importantly, coverage levels above 95\% offer no practical advantage in this dataset, as all minors are already correctly identified while interval widths become excessively large. At a coverage of 90\%, only 1 out of 38 minors had a minimum age exceeding 18 years, while the resulting prediction intervals remained substantially narrower, providing a more informative estimate for adults. These results suggest the practical utility of conformal prediction intervals for legally conservative age estimation without unnecessarily sacrificing precision.

\begin{table}[h!]
\footnotesize
\caption{Results for estimating the lower bound of a prediction interval at different coverage levels. PPV denotes the positive predictive value, where the positive class corresponds to individuals older than 18 years. For example, if an individual’s predicted age is 20.5 years, using an 80\% prediction interval yields a lower bound of 18.41 years, leading to classification as an adult. However, although the PPV at this level is 96\% (i.e., 96\% of individuals classified as adults are truly adults), an expert may consider this insufficient and instead prefer a higher coverage level (e.g., 95\%), which would result in classifying the individual as a minor.}
\label{tab:pred_intervals}
\begin{tabular*}{\linewidth}{@{\extracolsep{\fill}} cccccc }
\toprule
Coverage & d     & PPV   & Sensitivity & Specificity & Accuracy \\ \midrule
0\%      & 0     & 0.87 & 0.96       & 0.26       & 0.85    \\
60\%     & 1.44 & 0.96 & 0.82       & 0.82       & 0.82    \\
80\%     & 2.09 & 0.98 & 0.72       & 0.92       & 0.75    \\
90\%     & 2.97 & 0.99 & 0.62       & 0.97       & 0.68    \\
95\%     & 3.72 & 1     & 0.52       & 1           & 0.59    \\
99\%     & 5.60 & 1     & 0.15       & 1           & 0.29    \\ \bottomrule
\end{tabular*}
\end{table}

\section{Discussion}\label{sec:discussion}
\subsection{Analysis of the clavicle detection method}
The results obtained for clavicle detection demonstrate that the proposed combination of shape- and geometry-based features enables highly accurate discrimination between clavicles and other structures. Furthermore, unlike CNN-based detection or segmentation approaches, which currently constitute the dominant paradigm for this task, our method operates on connected components extracted from a threshold-based bone segmentation, yielding several practical and methodological advantages.

First, it enables a semi-automatic labeling strategy, drastically reducing annotation time. When working with 3D images, manual segmentation (pixel-wise annotation) requires several hours per individual, and manual detection (bounding-box annotation) still requires several minutes. Our approach reduces this effort to seconds: a human expert is shown all detected connected components, and just selects two of them as clavicles. This facilitates the construction of large, high-quality datasets with minimal expert intervention. Moreover, the labeled data could be further used to train CNN-based methods if required.

Second, the method provides voxel-perfect segmentation and detection. CNN-based segmentation methods may correctly identify a clavicle but fail locally, e.g. missing small regions or incorrectly segmenting boundaries. In contrast, because our approach operates on complete connected components, it either detects the entire bone or excludes it, avoiding partial detections.

Third, our method is robust to variations in CT acquisition protocols. The selected clavicular features are largely invariant across populations and scanning settings (e.g. they are the same in full-body and thoracic CTs), while CNN-based methods are sensitive to changes in slice thickness, field of view, or scanner manufacturers, among others. As a result, the proposed approach is expected to generalize more reliably to unseen datasets.

Although our method generates accurate bone segmentations, we deliberately chose not to use the segmentation mask directly as input to our model. The reason behind this is that LAE from the MCE relies critically on the appearance of the epiphyseal gap and surrounding cartilaginous tissue, which often exhibits Hounsfield Unit values close to those of bone. Therefore, threshold-based segmentation can erroneously fill this gap, removing the information required for LAE. Instead, we use the segmented region to localize a region of interest in the original CT, preserving subtle patterns relevant for LAE.

\subsection{Comparison with state of the art}

To the best of our knowledge, only two previous studies explicitly addressed AI-based LAE using CT scans of the MCE \citep{qiu2024machine, wesp2024radiological}, with an additional work focusing on chronological age estimation rather than LAE \citep{sun2025chronological}. \autoref{tab:SOA_compare} shows a summary comparing all results.

\begin{table*}[h!]
\centering
\footnotesize
\caption{Comparison between the proposed method and previously published state-of-the-art approaches. MAE (in years) is reported for two settings: (i) the original test sets as reported in each respective publication, and (ii) our validation set, where representative implementations of all methods were trained and evaluated under a unified experimental protocol for model selection.}
\label{tab:SOA_compare}
\begin{tabular}{@{\extracolsep{\fill}} cccc }
\toprule
Method                               & \begin{tabular}[c]{@{}c@{}}Test set age range\end{tabular}      & \begin{tabular}[c]{@{}c@{}}Reported MAE in test\end{tabular} & \begin{tabular}[c]{@{}c@{}}Validation MAE (our dataset)\end{tabular} \\ \midrule
Human expert \citep{qiu2024machine}             & 14-30          & 1.94                                                           & -                                                                      \\
2 experts + SVM* \citep{qiu2024machine} & 14-30          & 1.73                                                           & -                                                                      \\

Mean of experts \citep{wesp2024radiological}  & 15-30          & 1.84                                                           & -                                                                      \\
Stacked 2D CNN \citep{qiu2024machine}                      & 14-30          & 1.77                                                           & 2.30                                                                   \\
3D CNN    \citep{wesp2024radiological}                           & 15-30          & 1.65                                                           & 1.75                                                                   \\
\textbf{Our method}                  & \textbf{14-26} & \textbf{1.55}                                                  & \textbf{1.34}                                                          \\ \bottomrule
\end{tabular}
\end{table*}

In \citet{qiu2024machine}, several approaches were evaluated, including linear regression on expert-assigned Kellinghaus stages (MAE 1.94 years), SVM-based fusion of stages from two experts (MAE 1.73 years), and a CNN trained on stacked 2D slices following UNet-based MCE segmentation (MAE 1.77 years). In \citet{wesp2024radiological}, the mean of expert-assigned stages yielded a MAE of 1.97 years, while an ensemble of 20 3D CNNs achieved a MAE of 1.65 years. All reported values correspond to test-set performance. Lastly, although not directly focused on LAE, in \citet{sun2025chronological} the age estimates obtained by a 3D CNN were used to compute the Kellinghaus stage \citep{kellinghaus2010enhanced}, reaching a cross-validation accuracy of 90\% when only considering individuals ranged 15 to 30 years old.

The proposed method achieved a test MAE of 1.55 $\pm$ 0.16 years, which can be considered a very competitive result compared to the state of the art \citep{qiu2024machine, wesp2024radiological}. However, direct numerical comparison must be interpreted with caution, as the cited works used different age ranges (14–30 in \citet{qiu2024machine} and 15–30 years in \citet{wesp2024radiological}) and different age distributions. Similarly, comparison with \citet{sun2025chronological} is not straightforward due to differences in task definition, metrics, and evaluation protocols.

Nevertheless, we experimentally evaluated representative alternatives inspired by these works under our own experimental setting. A 3D ResNet50 \citep{resnet} (\citet{wesp2024radiological} used ResNet18 as backbone, but ResNet50 yielded a MAE 0.30 years lower than ResNet18 in our case) trained directly on 3D MCE volumes, reflecting the works proposed by \citet{wesp2024radiological} and  \citet{sun2025chronological}, achieved a validation MAE of 1.75 years, higher than the 1.34 years obtained by the selected multi-slice approach. Likewise, stacking multiple 2D slices as channels in a 2D CNN, as proposed by \citet{qiu2024machine}, resulted in poor generalization, where the model was unable to learn significant patterns (validation MAE 2.30 years). These results provide empirical support for the effectiveness of the proposed multi-slice, multi-view formulation over both 3D CNNs and stacked 2D approaches when data availability is limited. 

To determine whether the performance difference between our method and the 3D CNN baseline is statistically significant, we evaluated the final selected versions of both models on the same held-out test set and performed paired statistical comparisons on the per-sample absolute errors. For each test sample, the absolute error of our method was compared with that of the baseline model. We computed the mean paired difference in absolute error (ours minus baseline) and estimated its 95\% confidence interval using nonparametric bootstrap with 10,000 resamples. In addition, a two-sided paired permutation (sign-flip) test \citep{good2005permutation} with 10,000 permutations was performed to obtain a p-value. Effect size was quantified using the paired Cohen's $d$ \citep{cohen2013statistical} (mean difference divided by the standard deviation of the paired differences), and the Hodges–Lehmann estimator \citep{hodges1963estimates} was reported as a robust measure of the median paired difference. All inferential analyses were conducted on per-sample absolute errors from the held-out test set.

The mean paired difference in absolute error was -0.46 years (95\% CI: -0.77 to -0.14 years), indicating that, on average, our method reduced the absolute error by approximately 0.46 years compared to the 3D CNN baseline. The bootstrap confidence interval did not include zero, suggesting a statistically reliable improvement. The paired permutation test confirmed this result ($p = 0.0078$), demonstrating that the observed performance gain is unlikely to be due to chance. The paired Cohen’s $d$ was -0.36, corresponding to a small-to-moderate effect size in favor of the proposed method. The Hodges-Lehmann estimator yielded a median paired difference of -0.47 years, consistent with the mean-based analysis and further supporting the robustness of the improvement. Overall, these results provide statistical evidence that the proposed multi-slice approach achieves significantly lower absolute error than the 3D CNN baseline on unseen test data.

With respect to 3D CNNs, the proposed method represents anatomical information more efficiently by incorporating domain knowledge into the data representation itself. Rather than learning spatial relevance from full volumetric inputs, the model is explicitly guided toward anatomically informative slices, reducing the dimensionality of the learning problem. This results in a model that is computationally more efficient during both training and inference, and less prone to overfitting, particularly in limited-data scenarios.

In the case of stacked 2D CNNs, where multiple slices are treated as input channels of a single network, the proposed multi-branch design preserves slice-level feature extraction before fusion. In stacked 2D approaches, the network must learn cross-slice interactions from the first convolutional layer, potentially diluting slice-specific patterns and increasing the difficulty of optimization. In contrast, our architecture allows each slice to be processed independently through a dedicated backbone, ensuring robust feature learning at the slice level before higher-level aggregation. This structured fusion strategy better aligns with the anatomical interpretation process and leads to improved generalization performance.

\subsection{Analysis of the age estimation results}

A central finding of this study is that data processing and anatomical guidance are substantially more important than architectural complexity when estimating age from 3D medical images. Starting from a 3D CNN baseline with a validation MAE of 1.75 years, we progressively improved performance through informed simplifications. Using only five axial slices yielded a MAE of 1.78 years, which was reduced to 1.66 years just by improving slice selection. Adding the proposed data augmentation lowered it to 1.49 years, incorporating multiple views led to a validation MAE of 1.43 years. Lastly, using both clavicles to estimate age further reduced it to 1.34 years.

This progressive improvement was achieved not by increasing model capacity, but by guiding the model towards anatomically relevant information. The results indicate that the available dataset is insufficient for a model to autonomously learn the relevant spatial patterns from raw 3D volumes. Instead, explicitly focusing the network on the most informative slices enables more efficient learning and better generalization.

Several conclusions follow from the ablation study: (i) age-discriminant information is concentrated in a small number of slices around the most relevant anatomical location; (ii) coronal and axial views contain complementary information, while sagittal slices contribute little to performance; (iii) two slices per relevant view are sufficient, and adding more does not improve accuracy; and (iv) the effectiveness of the proposed data augmentation strategy suggests that slight imprecision in slice selection does not substantially degrade performance, provided that selected slices remain close to the most relevant region.

These findings closely align with forensic practice, where experts focus on specific axial and coronal slices rather than volumetric information \citep{wittschieber2014value, garamendi2023estimacion, el2015age}. In this regard, the proposed multi-slice approach mirrors expert reasoning more closely than end-to-end 3D CNNs, while also offering improved interpretability through slice-level analysis.

Regarding test-set performance, the proposed method achieved a MAE of 1.55 $\pm$ 0.16 years, with nearly identical errors for males (1.56 $\pm$ 0.19 years) and females (1.52 $\pm$ 0.35 years), indicating no detectable sex-related performance bias. This finding is consistent with recent studies reporting comparable accuracy across sexes in AI-based clavicle age estimation \citep{qiu2024machine}. However, it should be noted that males constitute 73\% of our dataset. While this represents an imbalanced distribution, it reflects real-world forensic practice, where the majority of individuals undergoing legal age assessment are male. In this sense, the cohort composition is aligned with the intended operational scenario of the method. Nevertheless, and although no performance bias is observed, the predominance of male subjects may limit the ability to detect more subtle sex-specific effects. Future studies including larger and more balanced cohorts would therefore be valuable to further assess robustness across different demographic settings and populations.

Age-stratified analysis showed particularly low errors between 17 and 25 years, the most relevant range for LAE. Higher errors observed below this range are not problematic in practice, as hand radiographs and dental methods are more precise and are applied earlier according to established forensic protocols \citep{schmeling2016forensic, cunha2009problem}. Errors above 25 years are unlikely to have practical legal implications, as individuals are far from the legal threshold of 18 years, and even large underestimations have no ethical or legal consequences \citep{garamendi2005reliability}.

\subsection{Explainability}

\begin{figure}[h!]
    \centering
    \subfloat{
        \includegraphics[width=\linewidth]{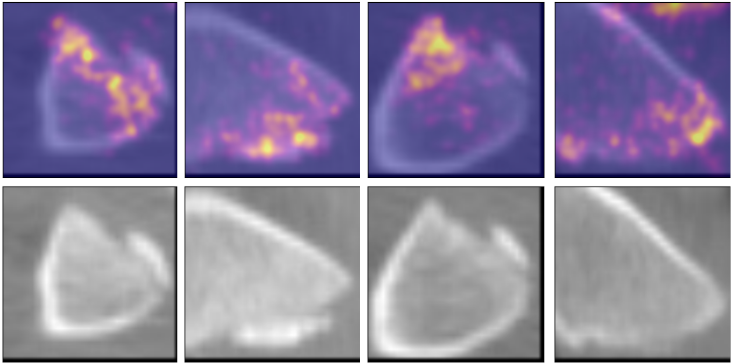}
    }
    
    \subfloat{
        \includegraphics[width=\linewidth]{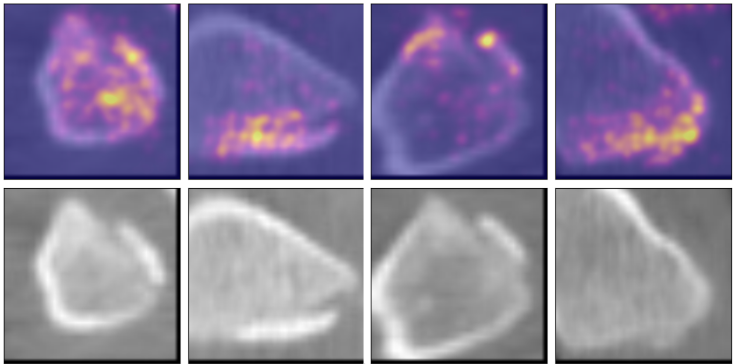}
    }
    \caption{Example of the activation maps obtained for the right (estimated as 18.78 years old, two first rows) and the flipped left (estimated as 18.38 years old, third and fourth row) clavicles of an individual from the test set. They correspond to the axial view (first and third column) and to the coronal view (second and fourth column). Both the activations and the used images are shown. The individual is 18.583 years old, and was estimated as 18.579 years old (mean of both clavicles). Three decimal places are used, instead of the two decimals employed in the rest of the paper, in order to better observe the minimal differences between the real and the predicted age. The more yellow a pixel, the more weight it has on the estimation.}
    \label{fig:clavs_exp}
\end{figure}

As shown in \autoref{fig:clavs_exp}, IG analysis indicates that the model consistently focuses on the medial epiphysis and adjacent diaphysis, particularly on the appearance and definition of the epiphyseal scar. Well-defined cortical borders and ossification patterns received high attribution, whereas poorly ossified regions exhibited more diffuse activations. These features correspond closely to the criteria used by human experts when applying the Kellinghaus staging system \citep{schmeling2016forensic, kellinghaus2010enhanced}. This alignment between learned features and established forensic criteria suggests that the model relies on clinically meaningful patterns, thereby supporting the interpretability of the proposed approach.

\subsection{Applicability in real forensic casework}

The proposed framework is designed with real forensic applicability in mind. First, the clavicle detection approach is generalizable to different CT acquisition protocols, including variations in scanned anatomical regions and fields of view, provided that the clavicle is visible. Unlike 3D CNN-based detection methods that operate directly on full volumes and may be sensitive to positional variability or scale differences, our approach relies on connected-component analysis in real physical space. This allows clavicle detection to be performed independently of its relative position within the scan.

Once detected, each clavicle is standardized by extracting a fixed-size volume centered on the MCE, defined as a bounding box of $2.5\times2.5\times2.5$ cm in real-world units. This spatial normalization ensures that subsequent processing is independent of scanner resolution or acquisition scale. The standardized volume is then processed by the slice-selection and LAE CNNs in a fully automated manner, enabling a consistent and reproducible pipeline.

Furthermore, as far as we know, this is the first MCE-based AI method for LAE that incorporates conformal prediction to generate calibrated prediction intervals. These intervals allow direct implementation of the minimum age concept recommended in European forensic protocols \citep{schmeling2016forensic}. Rather than providing only a point estimate, the method yields a statistically grounded lower bound, facilitating legally conservative decision-making and straightforward integration into multi-factorial age assessment procedures.

Finally, the framework can be deployed without demanding computational requirements. In contrast to volumetric 3D CNNs, which require substantial GPU memory and longer inference times, the proposed multi-slice architecture operates on a limited number of 2D slices. As a result, inference is fast and can be performed on standard workstation hardware in less than 20 seconds, where the major time bottleneck is computing shape descriptors for all bones within the clavicle detection process. This makes the method suitable for integration into operational forensic software environments. In fact, the system is currently being incorporated into the Skeleton-ID platform \citep{skeletonid}, enabling practical application in real forensic casework.

\subsection{Limitations}

The present study is based on post-mortem CT scans from the NMDID \citep{edgar2020new}, which differ in several respects from clinical CT acquisitions of living subjects, including patient positioning, soft-tissue characteristics, motion artifacts, and acquisition protocols. Although the clavicular ossification patterns targeted by the proposed method are primarily determined by bone morphology, these factors may nonetheless affect image appearance and, consequently, model performance. Therefore, the reported results should be interpreted as reflecting performance on post-mortem imaging, and external validation on clinical CT data from living subjects will be an important step before deployment in real-world forensic workflows.

\section{Conclusions and future work}
This work presents a complete AI-based framework for LAE from clavicle CT images. The proposed pipeline integrates: (i) robust clavicle and MCE detection based on connected component analysis, (ii) anatomically informed slice selection guided by IG; (iii) a multi-slice, multi-view CNN for age estimation; and (iv) computation of a prediction interval using conformal prediction. The method achieved a test MAE of 1.55 $\pm$ 0.16 years, outperforming alternative state-of-the-art 3D and stacked 2D approaches evaluated under the same experimental conditions.

To the best of our knowledge, this is the first study to incorporate conformal prediction intervals into AI-based clavicle LAE, enabling principled application of the minimum age concept in accordance with forensic guidelines \citep{schmeling2016forensic}. This facilitates future deployment in forensic practice, where legally conservative decisions are mandatory.

Beyond LAE, the proposed clavicle detection method constitutes a general-purpose tool for anatomical localization in CT images. Its extension to other skeletal structures, either directly or via morphological operations, represents a promising direction for future research and could substantially reduce preprocessing effort in a wide range of medical imaging applications. Moreover, the method can improve detection or segmentation performance in limited-data settings.

Finally, the proposed approach is designed to integrate seamlessly into a multi-factorial forensic LAE framework, alongside hand radiographs and dental imaging, as recommended by European protocols  \citep{schmeling2016forensic}. Together with our previously published work on dental age estimation \citep{venema2025trustworthy} and ongoing work on hand X-ray analysis, this study provides a key component toward a unified, trustworthy, and legally compliant forensic LAE system. Future work will focus on integrating the developed method into this multi-factorial framework and validating the complete system in both simulated and real-world forensic LAE scenarios.

\section*{Supplementary information}

The identifiers of the specific cases from the New Mexico Decedent Image Database that we used to train, validate, and test the developed method are provided as supplementary material. The dataset can be accessed at \url{https://nmdid.unm.edu/welcome}.

\section*{Declarations}

\subsection*{Competing interests}
The authors declare the following financial interests/personal relationships which may be considered as potential competing interests: Oscar Ibáñez and Pablo Mesejo report a relationship with Panacea Cooperative Research S. Coop. that includes board membership. All authors have the patent AGE ESTIMATION METHOD AND SYSTEM BASED ON DATA AGGREGATION pending to Panacea Cooperative Research S. Coop., University of A Coruña, and University of Granada.

\subsection*{Funding}

This publication is part of the R\&D\&I project PID2024-156434NB-I00 (CONFIA2), funded by \\ MICIU/AEI/10.13039/501100011033 and ERDF/EU, and has received funding from the European Union’s Horizon 2020 research and innovation programme under the Marie Skłodowska-Curie grant agreement No 101026482 —  UMAFAE — H2020-MSCA-IF-2020 and by Spanish Red.es under grant Skeleton-ID2.0 (2021/C005/00141299). Mr. Venema's work is funded by grant number 09/942572.9/23 within the 2023 call for aid from the Community of Madrid to finance projects that have obtained a Seal of Excellence within the European Innovation Council’s Accelerator Program. Dr. Ibañez’s work is funded by the Spanish Ministry of Science, Innovation and Universities under grant RYC2020-029454-I and by Xunta de Galicia, Spain by grant ED431F 2022/21. Additionally, some research included in this workshop has been funded by CITIC, which as a center accredited for excellence within the Galician University System and a member of the CIGUS Network, receives subsidies from the Department of Education, Science, Universities, and Vocational Training of the Xunta de Galicia. Additionally, it is co-financed by the EU through the FEDER Galicia 2021-27 operational program (Ref. ED431G 2023/01).


{\footnotesize
\bibliographystyle{elsarticle-harv}
\bibliography{refs}
}


\end{document}